\documentclass[10pt, a4paper]{article}

\usepackage[final]{lrec2026} % this is the new style
\usepackage{microtype}
\usepackage{booktabs}
\usepackage{adjustbox}
\usepackage{multirow}
\usepackage{amsmath}
\usepackage{framed}
\usepackage{subcaption}
\usepackage{dirtytalk}

\title{A Diagnostic Benchmark for Sweden-Related Factual Knowledge}

\name{Jenny Kunz} 

\address{Department of Computer and Information Science \\
Linköping University \\
         jenny.kunz@liu.se\\}

\abstract{
  Many Swedish benchmarks are translated US-centric benchmarks, and therefore not suitable for testing knowledge that is particularly relevant, or even specific, to Sweden. We therefore introduce a manually written question-answering benchmark specifically targeted to Sweden-related personalities and events, many of which receive very limited coverage in international media. Our annotators drew inspiration from a popular radio program featuring public figures from culture and media, as well as major sports events in Sweden. The dataset can be used to measure factual recall across models of varying sizes and degrees of Swedish coverage, and allows to probe cross-lingual factual consistency as it contains English translations. 
Using the dataset, we find that smaller models with stronger Swedish coverage perform comparably to a three times larger multilingual model in recalling Sweden-related facts. We also observe that continued pre-training on Swedish generally improves factual knowledge but also leads to forgetting of a part of the previously known information. These results demonstrate the dataset’s potential as a diagnostic tool for studying language adaptation and knowledge retention in multilingual models and during language adaptation. 
 \\ \newline  }

\begin{document}

\maketitleabstract

\section{Introduction}

LLMs are predominantly trained on English data and often struggle to transfer knowledge acquired in one language to others \citep{chua2025crosslingualcapabilitiesknowledgebarriers}. As a result, they are factually inconsistent across languages \citep{qi-etal-2023-cross, wang-etal-2025-lost-multilinguality}. Models that are instruction tuned on translated data get strong results when the evaluation data is translated from English, but perform less well on datasets originally annotated in the target language \citep{chen2024gooddatamultilingualinstruction}. \citet{chen2024gooddatamultilingualinstruction} hypothesize that this performance gap is due to a knowledge mismatch: the model’s mostly English-based knowledge may not align with the knowledge required to solve tasks in the target language. This shows that the common practice of translating evaluation datasets has fundamental limitations. 

For Swedish, most existing question answering (QA) datasets are either translated from English \citep{longpre-etal-2021-mkqa, singh-etal-2025-global, nielsen2024encodervsdecodercomparative} or generated by LLMs \citep{smart2025multiwikiqareadingcomprehensionbenchmark}. 
% \citep{hertzberg-lokrantz-2024-medqa} based on exams for medical licensing. 
While these are scalable ways to cover many languages, having \textit{only} such benchmarks leaves a gap in evaluating language-specific knowledge. Many multilingual datasets are US-centric and not equally meaningful in other cultures, and translation errors can affect the evaluation \citep{barth2025multilingualeuropeanlanguagemodels}. While there exist datasets that are derived from Swedish exams \citep{kurfali-etal-2025-swesat, hertzberg-lokrantz-2024-medqa}, there is a lack of datasets targeting \textit{facts closely tied to Sweden}, where much of the relevant information exists in Swedish. To address this gap, we introduce a dataset of 1,293 questions that capture local, culturally grounded knowledge about Sweden-related personalities and events. 

\begin{figure}
      \begin{framed}
      \textbf{Question:} \textit{What is the brand of the bike that Stig Johansson rides at Vätternrundan?} \\
      %\vspace{1mm}
      \textbf{Answer:} \textit{Husqvarna}.\\
      ------------------------------------------------------------ \\
      \textbf{Question: } \textit{In which singing competition did Toussaint \say{Tusse} Chiza represent Sweden?}\\
      \textbf{Answer: } \textit{Eurovision}.
    \end{framed}
    \caption{Two examples for question-answer pairs in our dataset: The first question is niche, while the answer to the second question is widely known. }
    \label{fig:example}
\end{figure}

Our questions are diverse in difficulty. 
Many of the questions in our dataset are niche even in Sweden (such as the first question in Figure~\ref{fig:example}), while others even receive some international coverage (such as the second question in Figure~\ref{fig:example}). 
In any case, we make sure that question is answerable through a quick Google search in Swedish. The samples are provided in both Swedish and English to enable direct comparison of model performance and factual consistency across the two languages. 

We use the new dataset to evaluate models of varying sizes and degrees of Swedish coverage. Our results show that smaller models with stronger Swedish representation (8B, 9B) perform comparably to a much larger multilingual model (27B) in factual knowledge about Sweden. Moreover, a model continually pre-trained on Swedish (8B) gains substantial new knowledge, but also forgets facts about Swedish domains that its base model previously knew. These findings suggest that our dataset can serve as a useful probe for studying language adaptation, helping to identify strategies for expanding a model's factual knowledge in a new language while minimizing catastrophic forgetting.

\section{Dataset Description}
\label{sec:dataset_description}

Our dataset consists of two subsets. 
The first subset, \textbf{Sommarpratare}, contains the majority of questions, 1,190 in total. \textit{Sommarpratare} refers to the hosts of \textit{Sommar i P1}, a long-running Swedish radio program broadcast. In each episode, a guest (typically a public figure, artist, athlete, or researcher) hosts the show and shares personal stories, reflections, and music selections. For this subset, each annotator was assigned a range of years and asked to create approximately three questions per speaker, with an emphasis on diversity in questions. The dataset covers the years 2018–2024.
The second subset, 
\textbf{Sports Events}, contains 102 questions. The largest subset focuses on \textit{En Svensk Klassiker}, a popular collection of long-distance races in four disciplines with 60 questions distributed across its constituent events: \textit{Vasaloppet} and \textit{Engelbrektsloppet} (skiing; 20 questions), \textit{Lidingöloppet} (running; 15 questions), \textit{Vätternrundan} (cycling; 15 questions), and \textit{Vansbrosimningen} (swimming; 10 questions). Among these events, Vasaloppet receives the broadest media coverage outside of Sweden, whereas the remaining events are assumed to have more limited international visibility. 
We collected additional sports-related questions through web searches, publicly available event listings (e.g., \url{lopplistan.se}), and contributions from colleagues. These questions span a broad range of sports, including further endurance sports events (e.g., marathons), team sports (e.g., football/soccer and handball), as well as horse racing, orienteering, and sailing. 

\paragraph{Annotation Process}
Questions were independently written by two student assistant annotators. Each question was verified by the other annotator. When formulating questions, annotators were instructed to consult reliable, high-quality sources such as Wikipedia, official event websites, or reputable news outlets. For each person or event, annotators used web searches to gather information before composing the question and answer. In the verification step, questions and answers were modified for clarity (e.g., adding missing information) and brevity, or discarded if considered unsuitable. 

\paragraph{Features}
Each \textbf{question} targets a specific factual detail, such as the name of an entity or a numerical value. When multiple units of measurement are possible (e.g., \textit{kilometers} versus \textit{miles} versus the Swedish \textit{mil}), the question explicitly specifies the unit corresponding to the gold-standard answer to ensure consistency.
\textbf{Answers} are edited to be minimal in order to make string-based evaluation metrics (particularly recall) as reliable approximators of model performance as possible.
For questions concerning recent events, some answers may depend on information available only after a specific point in time, which may lay behind the \textbf{cutoff date} for certain models. When relevant (i.e., the cutoff date is after 2021), we include this date or an estimated cutoff. 
A human \textbf{translation} of both the question and its corresponding answer is provided in English to support cross-lingual evaluation. 

\paragraph{Answer Type Distribution}

To get statistics about the types of the answers asked for in the dataset, we classify each answer into eight categories using spaCy's~\citep{honnibal2020spacy} named entity recognition (NER) pipeline on the English answers, with part-of-speech (POS)-tag-based fallback for answers not recognized as entities. The answer is first processed in the context of its question for better entity recognition; if no entity is detected, the dominant POS tag determines the category. The categories are: \textit{Person}~(NER: \texttt{PERSON}), \textit{Location}~(NER: \texttt{GPE}, \texttt{LOC}, \texttt{FAC}), \textit{Organisation}~(NER: \texttt{ORG}), \textit{Other Entity}~(NER: \texttt{NORP}, \texttt{EVENT}, \texttt{WORK\_OF\_ART}, \texttt{PRODUCT}, \texttt{LAW}, \texttt{LANGUAGE}; POS: \texttt{PROPN}), \textit{Numeric}~(NER: \texttt{CARDINAL}, \texttt{ORDINAL}, \texttt{QUANTITY}, \texttt{MONEY}, \texttt{PERCENT}; POS: \texttt{NUM}), \textit{Date/Time}~(NER: \texttt{DATE}, \texttt{TIME}), \textit{Concept}~(POS: \texttt{NOUN}, \texttt{ADJ}, \texttt{VERB}, \texttt{ADV}), and \textit{Other}.
\begin{figure}
    \centering
    \includegraphics[width=\linewidth]{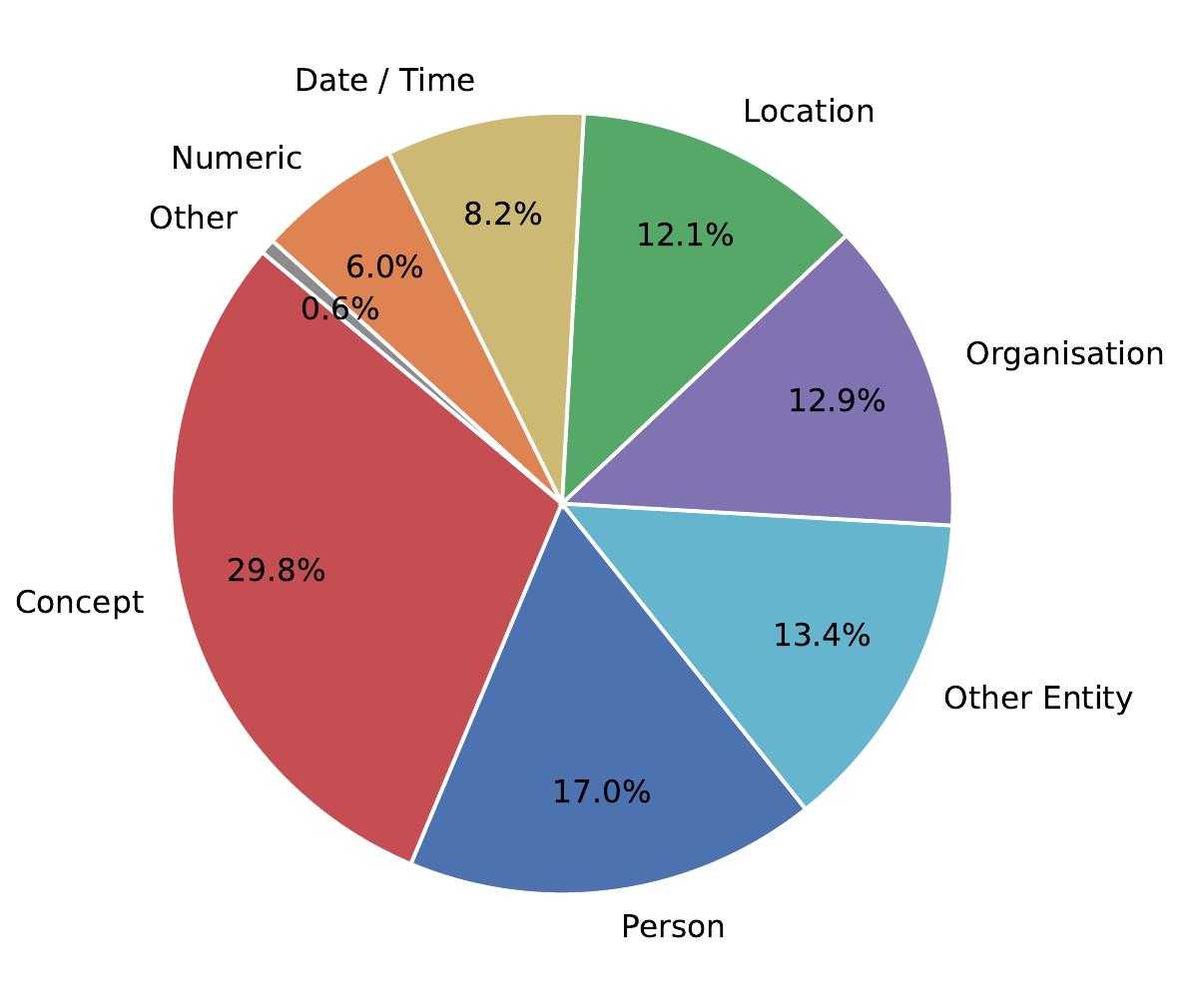}
    \caption{Distribution of answer types in the dataset, classified using spaCy NER and POS tagging. }
    \label{fig:answer_types}
\end{figure}
The results are in Figure~\ref{fig:answer_types}. The largest category is \textit{Concept} (385, 29.8\%), followed by \textit{Person} (220, 17.0\%), \textit{Other Entity} (173, 13.4\%), \textit{Organisation} (167, 12.9\%), and \textit{Location} (156, 12.1\%). \textit{Date/Time} and \textit{Numeric} answers account for 8.2\% and 6.0\%, respectively. 

\paragraph{Release} Our dataset is openly available at  \url{https://huggingface.co/datasets/liu-nlp/swedish-facts-v1}. 

\section{Experimental Setup}

\paragraph{Models} 
We evaluate two families of open-weight multilingual models on our dataset: Gemma-3 (270M, 1B, 4B, 12B, 27B) \citep{gemma_2025} and EuroLLM (1.7B, 9B) \citep{martins2024eurollmmultilinguallanguagemodels}. 
In addition, we use our dataset to test whether continued pre-training contributes to language-specific knowledge by comparing AI Sweden's fine-tune of LLaMA-3\footnote{\url{https://my.ai.se/resurser/3070}}, which was further pre-trained on Swedish, Norwegian, and Danish text, to its base model, LLaMA-3 \citep{grattafiori2024llama3herdmodels}. 
All models are used in their instruction-tuned variant. We apply the chat template and set the temperature to 0. 

\paragraph{Prompts}
We prepend a brief instruction to each prompt: “\textit{Give a short answer to the following question.}” For Swedish question–answer pairs, the models are prompted in Swedish; for the English translations, they are prompted in English.\footnote{Note that the performance may improve using English even for Swedish questions, but this would be unnatural as we want to evaluate target-language performance (compare \citet{poelman-lhoneux-2025-roles}).}

\paragraph{Metrics}
As described in Section~\ref{sec:dataset_description}, the answers in our dataset were designed to be minimal, allowing recall to serve as a reasonable approximation of model performance without relying on less interpretable or reproducible methods such as BERTScore \citep{Zhang2020BERTScore} or LLM-as-a-judge \citep{NEURIPS2023_91f18a12}. To compute the scores, texts are lowercased and punctuation removed. 
We use three basic metrics: exact match (EM), token-level F1, and recall. In addition, we measure the recall-based cross-lingual consistency in two setups: \textit{en $\mid$ sv}, that is, how often the correct English answer is recalled given that the correct Swedish answer is recalled for the same sample, and vice versa, \textit{sv $\mid$ en}, how often the correct Swedish answer is recalled given a correct English answer.
%This can be considered a simplified version of the RankC metric proposed by \citet{qi-etal-2023-cross} where we only consider the single most likely candidate. 

\section{Results}

\begin{table*}[ht]
\centering
\begin{tabular}{lrrrrrrrr}
\toprule
\multirow{2}{*}{\textbf{Model}} & \multicolumn{3}{c}{\textbf{Performance sv}} & \multicolumn{3}{c}{\textbf{Performance en}} & \multicolumn{2}{c}{\textbf{Consistency}} \\
\cmidrule(lr){2-4} \cmidrule(lr){5-7} \cmidrule(lr){8-9}
 & EM & F1 & Recall  & EM & F1 & Recall & en $\mid$ sv & sv $\mid$ en  \\
\midrule
gemma-3-270m & 0.00 & 0.09 & 1.86 & 0.00 & 0.24 & 5.42 & 47.82 & 18.64 \\
gemma-3-1b & 0.00 & 0.41 & 5.88 & 0.00 & 0.85 & 11.38 & 49.28 & 25.56 \\
gemma-3-4b & 0.00 & 3.37 & 12.15 & 0.00 & 3.93 & 15.09& 51.59 & 41.53 \\  
gemma-3-12b & 0.15 & 8.82 & 19.04 & 2.79 & 9.43 & 20.36& 63.01 & 58.94 \\
gemma-3-27b & 0.00 & 9.85 & 25.00 & 1.86 & 9.02 & 25.15 & 68.42 & 68.00\\\midrule
EuroLLM 1.7B & 0.08 & 2.71 & 10.53 & 0.00 & 2.41 & 15.79 & 56.61 & 37.74 \\
EuroLLM 9B & 0.00 & 5.31 & 23.07  & 0.00 & 5.61 & 25.39 & 68.79 & 62.50 \\
\midrule
LLaMa-3-8B & 0.08 & 3.98 & 19.89 & 0.00 & 2.88 & 21.21 & 58.75 & 55.11  \\
ai.se LLaMa-3-8B & 0.00 & 4.75 & 23.14 & 0.00 & 2.98 & 13.00 & 43.81 & 77.97 \\\bottomrule
\end{tabular}
\caption{\textbf{Evaluation results for all models.} The first six columns show performance in each individual language. The last two columns report (recall-based) cross-lingual consistency: the proportion of correct English answers given correct Swedish answers (en $\mid$ sv), and vice versa (sv $\mid$ en). }
\label{tab:results}
\end{table*}

In Table \ref{tab:results}, we see that overall performance on our dataset is relatively low: no model answers more than a quarter of the questions correctly. However, accuracy steadily increases with model size, indicating that larger models recall more fine-grained facts about Swedish personalities and events.

Most models perform better when questions are asked in English, except for AI Sweden LLaMA, which consistently responds in Swedish and therefore receives a lower recall score due to the mismatched answer language. Gemma-27B performs similarly well across both languages. 

Interestingly, EuroLLM demonstrates strong knowledge of Swedish facts relative to its size: Results are comparable to the AI Sweden LLaMA, which covers far fewer languages, while also achieving much higher scores on the English questions than AI Sweden LLaMa. Its recall is similar to that of Gemma-27B, though its F1 score is lower, likely because (as we observe) Gemma-27B follows instructions more closely and tends to produce concise answers without additional context. 

\paragraph{Consistency}
The last two columns in Table~\ref{tab:results} show limited overlap between answers given in Swedish and English by the same model. Among the correct answers in one language, only up to 68--69\% are also correct in the other. These findings align with previous work showing that LLMs struggle to transfer knowledge across languages \citep{goldman2025eclekticnovelchallengeset, wang-etal-2025-lost-multilinguality}.
The only exception where the number is higher is the \textit{sv $\mid$ en} setup in AI Sweden Llama. This exception is however caused by this model exclusively answering in Swedish due to Swedish-only instruction tuning, resulting in very low English performance compared to its high Swedish accuracy. 

\paragraph{Continued Pre-Training}

We compare the models' knowledge of Swedish domains before and after continued pre-training (CPT) on Swedish data.
In the first setup, where the models are prompted in Swedish, 165 questions are answered correctly by both models. The original LLaMA answers 92 questions correctly that AI Sweden LLaMA misses, while the latter answers 134 correctly that the original does not. Overall, AI Sweden LLaMA retains 64.20\% of the questions correctly answered by the original model. Conversely, the original LLaMA answers correctly only 55.18\% of the questions that AI Sweden LLaMA gets right, indicating a substantial knowledge gain from CPT, though the dataset remains far from solved. However, CPT also causes some catastrophic forgetting: AI Sweden LLaMA fails on 92 of the 257 samples (35.80\%) that the original model previously answered correctly.

The new knowledge gained through CPT mainly concerns less internationally known people, such as Swedish TV presenters and actors. The losses are however unexpected: For example, when asked in Swedish where the Gothenburg half marathon (\textit{Göteborgsvarvet}) starts, the original LLaMA correctly answered \textit{Slottskogen} (a park in Gothenburg) while the AI Sweden model did not. Given that this race is one of the world’s largest annual running events and, with both English and Swedish online coverage, it is not unexpected that the English model already had the knowledge; however, coverage in Swedish-language training data should be higher given that the race certainly is much more famous in Sweden than abroad.

In the second setup, where both questions and answers are in English, the original LLaMA answers 190 questions correctly that AI Sweden LLaMA misses, while the latter answers 92 that the original does not. Only 72 questions are answered correctly by both. However, this comparison is less reliable, as the AI Sweden model consistently only responds in Swedish, leading to semantically correct answers being marked incorrect by our string-based metrics. 

\paragraph{What knowledge do models gain with size?}
As model size increases, scores improve consistently. Smaller models often produce hallucinated or overly generic answers, while larger ones recall concrete facts more accurately. For example, when asked which instrument Swedish jazz singer Svante Thuresson played before focusing on singing, Gemma-27B correctly answers \textit{drums}, whereas Gemma-4B incorrectly says \textit{saxophone}; a plausible but wrong guess. Similarly, for the question what sport participants in \textit{O-Ringen} compete in, Gemma-27B correctly answers \textit{orienteering}, while Gemma-4B hallucinates that it is a \textit{martial art and obstacle course} competition.

\paragraph{Problems with Answer Recall as a Metric}
Although recall is a simple and interpretable metric, using it as the main measure has important limitations. Beyond the issue of factually correct answers appearing in a different language, recall can also overestimate performance when models give partially correct or mixed answers. Any overlap between the model output and the gold answer may yield a high score even if the response contains factual errors.
For example, when asked (in Swedish) about Sven-Göran Eriksson’s occupation, Llama-3-8B responds (translated): \say{\textit{Sven-Göran Eriksson was a Swedish soccer coach. He was, among other things, the coach of the Swedish national team, Lazio, Roma, Manchester City, Leeds United, Aston Villa, and the German national team}.}
The first sentence correctly identifies his occupation (\say{\textit{soccer coach}}), but the continuation introduces inaccuracies about the teams he managed. Since recall only checks for the presence of the correct substring, this response is counted as correct. 
This illustrates that verbose models may seem more accurate than they truly are. 

\section{Related Work}

There are several QA datasets for Swedish, most of which are derived from English datasets or generated by LLMs. MKQA \citep{longpre-etal-2021-mkqa} is a multilingual open-domain QA dataset that explicitly excludes geographically dependent questions. ScandiQA \citep{nielsen-2023-scandeval} extends MKQA with supporting documents to turn it into a reading comprehension dataset. MMLU \citep{hendrycks2021measuring}, a multiple-choice dataset covering diverse subjects, was translated into Swedish as part of the Global MMLU project, which also removed US-specific content \citep{singh-etal-2025-global}. \citet{smart2025multiwikiqareadingcomprehensionbenchmark} LLM-generate a large multilingual QA dataset from Wikipedia texts in each language.
The Swedish university entrance exam (SweSAT) \citep{kurfali-etal-2025-swesat} assesses vocabulary, reading comprehension, mathematical problem solving, and logical reasoning but focuses less on factual knowledge. Relatedly, MedQA-SWE by \citet{hertzberg-lokrantz-2024-medqa} is based on exams for medical licensing. 

Other multilingual QA datasets test factual knowledge from structured sources using template-based questions, which are scalable because templates can be translated in a controlled setup. Examples include MLAMA \citep{kassner-etal-2021-multilingual}, X-FACTR \citep{jiang-etal-2020-x}, and BMLAMA \citep{qi-etal-2023-cross}. \citet{kassner-etal-2021-multilingual} show that mBERT’s performance depends strongly on the query language, while \citet{qi-etal-2023-cross} find that geographically related languages share some encoded knowledge. mParaRel \citep{fierro-sogaard-2022-factual} probes (inter-language) consistency, showing that models are less consistent in non-English languages. KLAR \citep{wang-etal-2025-lost-multilinguality} tests cross-lingual factual consistency and finds that, although facts are stored in a language-independent space, models still struggle to transfer them across languages. The authors propose bypassing final language-specific layers to improve factual transfer.
% The INCLUDE benchmark \citep{romanou2025include} compiles local exam questions in 44 languages.
Most related to our work, ECLEKTIC \citep{goldman2025eclekticnovelchallengeset} is a dataset for evaluating cross-lingual fact recall using LLM-generated questions and answers based on Wikipedia articles available in only one language. This offers a scalable way to test language-specific knowledge. Although ECLEKTIC does not include Swedish, it provides a promising direction for future work. Unlike their dataset, ours does not explicitly filter for language-\textit{exclusive} topics, so many entities will have international coverage.

\section{Conclusion}

We introduced a new manually curated question–answering dataset for Swedish targeting factual knowledge about Sweden-related personalities and events, as those domains will not be appropriately represented in translated benchmarks. Our dataset includes questions and (short) answers, as well as English translations to evaluate cross-lingual consistency. 
Using this dataset, we evaluated a range of multilingual and Swedish-adapted models and found that smaller models with stronger Swedish coverage (8B--9B) can perform on par with a much larger multilingual model (27B). Continued pre-training on Swedish substantially increases factual recall but also leads to catastrophic forgetting of knowledge that the base model previously possessed, showing limitations of such naive strategies to language adaptation. Our findings highlight the dataset's potential as a diagnostic probe for language adaptation, allowing researchers to systematically study how models acquire, retain, or lose relevant factual knowledge when adapted to Swedish. 
In future work, we envision using this benchmark to explore strategies that mitigate forgetting and enhance local knowledge during adaptation, such as varying proportions of multilingual data, language mixing strategies, replay, regularization methods, and architectural modifications. 

\section*{Limitations}
In this work, we focus on smaller open-weight models that can be compared across sizes or with or without continued pre-training on Swedish. We did not include the largest state-of-the-art LLMs and commercial systems.

Our dataset covers a relatively narrow range of domains, limited to public figures featured in a national radio program and major sports events. It does not specifically target information available \textit{only} in Swedish, and we currently lack annotations indicating how well these facts are covered in English sources. Such measures may be added in future versions of the dataset.

\section*{Acknowledgments}
Special thanks to our student assistant annotators Anja Jarochenko and Salome Kasendu for their contributions to this dataset, and to my colleagues in the LiU NLP group who contributed to the list of sports events. This research was supported by TrustLLM funded by Horizon Europe GA 101135671. The computations were enabled by the National Academic Infrastructure for Supercomputing in Sweden (NAISS), partially funded by the Swedish Research Council through grant agreement no. 2022-06725. 

\section{Bibliographical References}\label{sec:reference}

\bibliographystyle{lrec2026-natbib}
\bibliography{lrec2026-example}

%\section{Language Resource References}
%\label{lr:ref}
\bibliographystylelanguageresource{lrec2026-natbib}
%\bibliographylanguageresource{languageresource}

\end{document}